\documentclass[a4paper]{article}
\usepackage{INTERSPEECH2014,amssymb,amsmath,epsfig,url}
\ninept
\def\reg{{\rm\ooalign{\hfil
     \raise.07ex\hbox{\scriptsize R}\hfil\crcr\mathhexbox20D}}}

\title{A Deep Learning Approach to Data-driven Parameterizations for
  Statistical Parametric Speech Synthesis}

\makeatletter
\def\name#1{\gdef\@name{#1\\}}
\makeatother \name{{\em Prasanna Kumar Muthukumar, Alan W Black}}

\address{Carnegie Mellon University \\
  Pittsburgh, USA \\
{\small \tt pmuthuku@cs.cmu.edu, awb@cs.cmu.edu}}

\begin{document}
\maketitle
\begin{abstract}
Nearly all Statistical Parametric Speech Synthesizers today use Mel
Cepstral coefficients as the vocal tract parameterization of the
speech signal. Mel Cepstral coefficients were never intended to work
in a parametric speech synthesis framework, but as yet, there has been
little success in creating a better parameterization that is more
suited to synthesis. In this paper, we use deep learning algorithms to
investigate a data-driven parameterization technique that is designed
for the specific requirements of synthesis. We create an invertible,
low-dimensional, noise-robust encoding of the Mel Log Spectrum by
training a tapered Stacked Denoising Autoencoder (SDA). This SDA is
then unwrapped and used as the initialization for a Multi-Layer
Perceptron (MLP). The MLP is fine-tuned by training it to reconstruct
the input at the output layer. This MLP is then split down the middle
to form encoding and decoding networks.  These networks produce a
parameterization of the Mel Log Spectrum that is intended to better
fulfill the requirements of synthesis.  Results are reported for
experiments conducted using this resulting parameterization with the
ClusterGen speech synthesizer.
\end{abstract}
\noindent{\bf Index Terms}: Statistical Parametric Speech Synthesis,
Deep Learning, Parameterization

\section{Introduction}
The speech coder used in modern Statistical Parametric Speech
Synthesis\cite{zen2009statistical} has remained largely unchanged for
a number of years. The standard coding technique is usually a variant of Mel
Cepstral analysis\cite{tokuda1994mel}. While many different
parameterizations of the spectrum have been developed for synthesis\cite{dutoit1996use}\cite{stylianou2001applying}\cite{dutoit1993mbr}\cite{moulines1990pitch},
few have yet managed to survive in the long run. The most obvious
indications of this are the systems that are submitted to the annual
Blizzard Challenge\cite{Blizzard}. Very few statistical parametric systems submitted
to the challenge since its inception use vocoders that do not use Mel Cepstral
coefficients. Even highly successful techniques like the various flavors of
STRAIGHT\cite{kawahara2008tandem} are rarely used by the synthesizer
directly. These are usually converted into Mel Cepstral coefficients (MCEPs)
before being used by statistical parametrical systems. 

This lack of new parameterizations that perform better than
MCEPs is especially intriguing considering the amount of research
effort that has gone into finding a replacement.  An ideal
parameterization for statistical parametric synthesis will have to
fulfill all of the following requirements:
\begin{itemize}
  \item It must be invertible
  \item It must be robust to corruption by noise
  \item It must be of sufficiently low dimension
  \item It must be in an interpolable space
\end{itemize}
Even if a parameterization technique were invented that could comply with
three of the above four requirements, the technique would be useless
if it did not at least partially satisfy the remaining one. Therein
lies the difficulty of inventing a new parameterization. Mel Cepstral
coefficients satisfy all of these requirements to a
reasonable extent. However, this representation is not perfect and
places a major bottleneck on the naturalness of modern parametric
speech synthesizers. Techniques such as \cite{tokuda1995speech} and
\cite{tomoki2007speech} rectify some of the problems that occur with
this representation but the Mel Cepstral representation still leaves 
plenty of room for improvement.

In this age of big data and deep learning, it behooves us to try
to construct a parameterization purely from data which might be more
adept at dealing with all these constraints.

\section{Stacked Denoising Autoencoder}

\begin{figure}[htb!]
\centerline{\includegraphics[width=5.0cm]{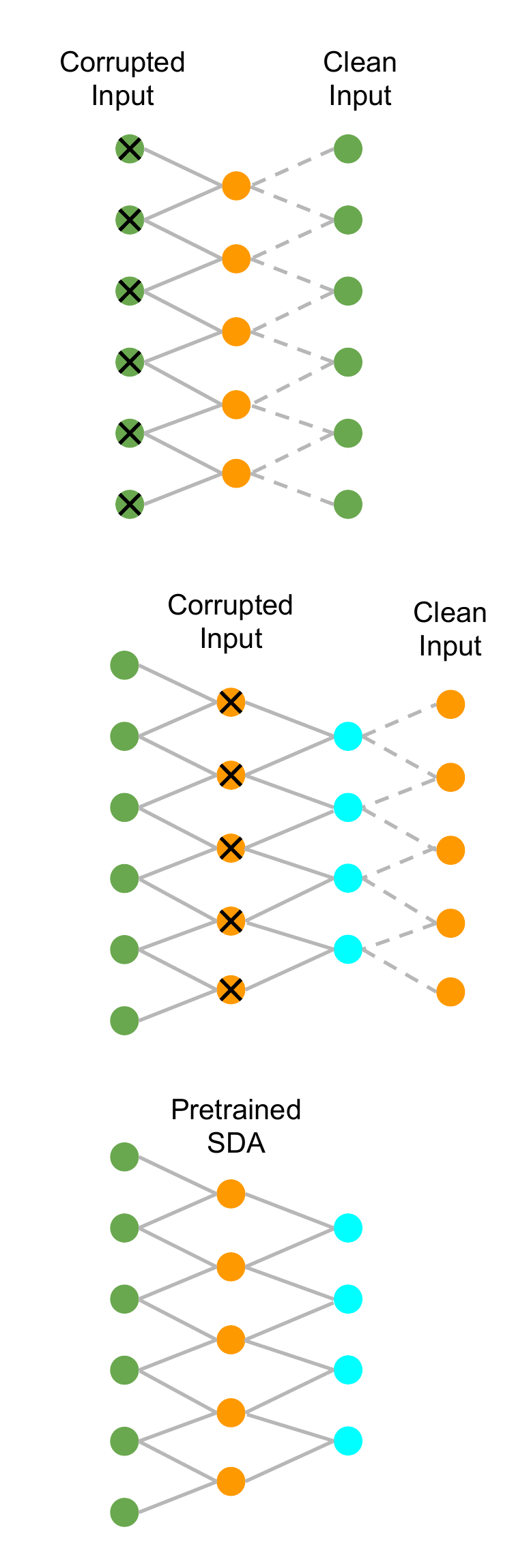}}
\caption{{\it SDA Pretraining}}
\label{pretraining}
\end{figure}

Neural networks themselves have existed for many years but the
training algorithms that had been used were incapable of effectively
training networks that had a large number of hidden
layers\cite{larochelle2009exploring}. This is
because the standard technique used for training a neural network is
the backpropagation algorithm\cite{rumelhart1988learning}. The
algorithm works by propagating the errors made by the neural network
at the output layer back to hidden layers and then adjusting the
weights of the hidden 
layers using gradient descent or other techniques to minimize this
error. When the network is very deep, the propagated error to the
first few hidden layers becomes very small. As a result, the
parameters of the first few layers change very little in training. One
strategy that was developed in recent years was to start off by
training the neural network one pair of layers at a time and then
building the next pair on top of previous
ones\cite{hinton2006fast}\cite{hinton2006reducing}. This step is
called 
\emph{pretraining} because the weights that are obtained through this
process are used as the initialization for the backpropagation
algorithm. Pretraining techniques are believed to provide an
initialization much closer to the global optimum compared to the
random initializations that were originally used.

Our search for a technique to create a purely data-driven
parameterization led us to the Stacked Denoising
Autoencoder (SDA) which was developed for pretraining deep neural
networks\cite{vincent2008extracting}. The
SDA is trained in a manner more or less identical to the
layer-wise pretraining procedure described in \cite{bengio2007greedy}
and \cite{hinton2006fast}. As the name suggests, the 
Stacked Denoising Autoencoder is constructed by stacking several
Denoising Autoencoders together to form a deep neural network. Each 
Denoising Autoencoder is a neural network that is trained such that
it reconstructs the correct input sequence from an
artificially corrupted version of the input provided to it. This
process is shown in Figure 
\ref{pretraining}. The network 
is fully connected between each layer but in the interest of clarity,
the figure will only show a limited number of connections. 

The SDA is of particular interest to parametric
speech synthesis because this network learns to
reconstruct a noisy version of the input from a lower dimensional set
of features. It is therefore apparent that the very definition of this
network fits the first three of the four requirements of an ideal
parameterization. We will discuss the fourth requirement in a later
section.

The SDA is actually rarely used in a task where the input needs to be
\emph{reconstructed} from the 
representation that the SDA transforms the input into. It is nearly always
used to provide a lower dimensional 
representation on top of which a classifier such as logistic
regression, or Support Vector Machines are used. An example of this is
the Deep Bottleneck Features that are used in Speech
Recognition\cite{hinton2012deep}\cite{gehring2013extracting}. However,
such approaches are less relevant to parametric synthesis since it is
\emph{not} a classification problem.

\section{Building Encoding and Decoding networks}

\begin{figure}[htb!]
\centerline{\includegraphics[width=5.5cm]{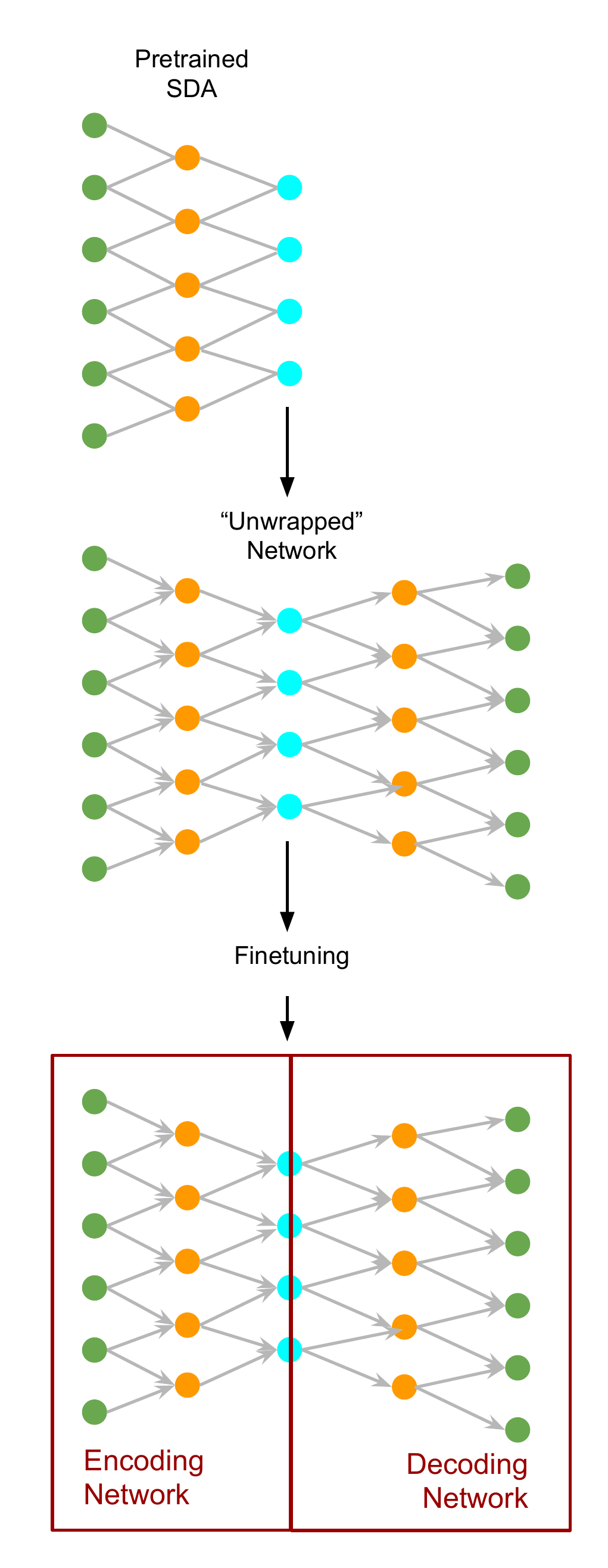}}
\caption{{\it Encoding and Decoding networks}}
\label{finetuning}
\end{figure}

The 'pretraining' process for our approach is identical to the one for
speech recognition. We build an SDA on our features by stacking
multiple Denoising Autoencoders that were built by learning to
reconstruct corrupted versions of the input. Once the SDA is trained,
we then \emph{unwrap} the SDA as shown in figure \ref{finetuning}. 

The unwrapped SDA acts as the initialization for a multilayer
perceptron (MLP). An $N$ layer SDA will produce an MLP with $2N-1$
layers. Backpropagation is
used to finetune the MLP such
that the output layer can reconstruct the input provided to the first
layer through the bottleneck in the middle. Once this finetuning has
been completed, this network is split down the middle into two
parts. The section from the input layer to the bottleneck region is
the \emph{encoding} network, while the section from the bottleneck
region to the output layer is the \emph{decoding} network. The
encoding network codes the speech signal into a representation which
is by design, invertible, robust to noise, and low dimensional. This
representation is the encoding that the synthesizer uses as the
parameterization of the speech signal i.e. it learns a mapping between
the phonemes of text and the values of this encoding. At synthesis
time, the synthesizer predicts values of this encoding based on the
input text. The decoding network converts this code back into a
representation of the speech signal. This approach is similar to the
one proposed for efficient speech coding in
\cite{deng2010binary}. Apart from the fact that \cite{deng2010binary}
proposes the use of the code for other applications, it is also
different in that it specifically looks for a binary encoding. Such binary
encodings are not very useful in a statistical synthesis framework
because binary representations are not interpolable while synthesis
is an inherently generative task.

\subsection{Input features}
In previous sections, we have discussed how a deep neural
network will build a low-dimensional noise-robust representation of
the speech signal, but what should our deep neural network actually
encode? To put it more explicitly, what should be the input to our
deep neural network that it can learn to reconstruct? Should it be the
actual speech signal itself, the magnitude spectrum, the complex
spectrum, or any of the other representations that signal processing
research has provided us? In theory, the input representation should
not matter since it has been proven that multilayer feedforward
networks are universal
approximators\cite{hornik1989multilayer}. However, this proof places
no constraints on the size or structure of the network. Nor does it
provide a training algorithm that reaches the global
optimum. Therefore, it is sensible to train the network on  a
representation that is known 
to be highly correlated with speech perception. Human hearing is known to
be logarithmic both in amplitude\cite{robinson1956re}, and
frequency\cite{stevens1937scale}. So, we propose that the Mel
Spectrum and Mel Log spectrum are the most suitable representations
that the network can be trained on. Despite using input and output layers
that were linear, the network had difficulty working with the wide
range of values in the Mel spectrum. Therefore, for the rest of this
paper, we will only describe our attempts at using a deep neural
network to get an invertible, low-dimensional, noise-robust
representation of the Mel Log Spectrum.

\section{Experiments and Results}
We built the SDAs and the MLPs using the Theano\cite{bergstraTheano}
python library, and the parametric Speech Synthesizer using
ClusterGen\cite{black2006clustergen}. The input to the neural network
was a 257-dimensional Mel Log Spectral vector which was obtained from
a 512-point FFT of a 25ms speech frame. The encoding obtained using
the network is 50-dimensional. This encoding size was chosen to make
it easier for us to compare the quality with the 50-dimensional MCEP
representation used in our baseline system. The Stacked Denoising
Autoencoder was built in a 257 x 125 x 75 x 50 configuration i.e. 257
nodes in the input layer, 125 in the first hidden layer, 75 in the
second, and 50 in the output layer. This
results in an MLP with a 257 x 125 x 75 x 50 x 75 x 125 x 257
configuration for fine-tuning. The encoding network will therefore
have a configuration of 257 x 125 x 75 x 50 and the decoding network,
50 x 75 x 125 x 257. In all of these networks, the layer that is
contact with the Mel Log Spectra is a linear layer with no non-linear
function involved. This is so that the layer can deal with the range
of values that the Mel Log Spectra can take. In all other layers, the
neurons have sigmoid activations. 

We ran experiments on 3 different voices: the RMS, and SLT voices
from the CMU Arctic databases\cite{kominek2004cmu}  and the Hindi
corpus released as part of the 2014 Blizzard
challenge\cite{Blizzard2014}. The intention 
was to test the setup across gender as well as across language. 

Evaluating the quality of the systems that we built poses an
interesting problem. The standard objective metric used in nearly all
evaluations of parametric speech synthesis is Mel Cepstral
Distortion\cite{kubichek1993mel}. However, this metric is likely inherently
unfair to our technique. This is because the default system that we
compare against, like most statistical parametric synthesizers, works
directly with the MCEPs. These systems optimize for the
root-mean-squared error of MCEP prediction. In other words, they
directly optimize for 
the metric. The technique that we are proposing gives us an
encoding which is optimized for parametric synthesis. But optimizing
for the prediction of this encoding need not necessarily optimize the
Mel Cepstral Distortion directly. Therefore, any MCD-based results
presented in this paper must be taken with a pinch of salt.

While our synthesizer might not directly optimize for MCD, the MCD
is nevertheless a good indicator of listener perception; the argument
here being that natural-sounding speech should have natural-appearing Mel
Cepstral parameters. So, we will
measure the quality of synthesis using the Mel Cepstra obtained from
the Mel Log Spectra of the decoding network. 

The first test is a simple analysis-resynthesis test. We measure how
well our learned encoding is able to reconstruct the Mel Log Spectra
of held-out test data. The results are shown in table \ref{Ana-Res}

\begin{table}[t,h]
\caption{\label{Ana-Res}\it{Analysis-Resynthesis}}
\centerline{
\begin{tabular}{ |c | c | } \hline
  Voice & MCD Scores \\\hline
  ARCTIC RMS & 4.354 \\
  ARCTIC SLT & 4.315\\ 
  Hindi corpus & 3.916 \\ \hline
\end{tabular}}
\end{table}

In all of the three cases, the deep neural network trained for the
voice is able to reconstruct the test set with relatively low
error. It is however not obvious what these numbers should be compared
against. 

One major hinderance in the use of deep neural networks is the amount
of effort that needs to go into the tuning of hyperparameters (batch
size, learning rate, number of epochs, etc\ldots). We did not want to
use a technique that would require a lot of tuning for \emph{each} new
speaker. So, all neural network settings were tuned for one speaker (RMS)
and identical configurations were used for the others. All the SDAs
were pretrained with a batch size of 20 for 50 epochs, and the MLPs
were trained with a batch size of 100 for 100 epochs. Based on the
results shown in table \ref{Ana-Res}, it appears that identical
hyperparameter settings work well for multiple speakers. 

As a test, we also tried using the RMS encoding and decoding networks
for Analysis-Resynthesis of SLT held out data. This resulted in an MCD
of 6.473, implying that the encoding that is learned by the deep
neural network is speaker specific. 

\subsection{Synthesis tests}

The next set of tests were on using the deep neural network's
50-dimensional encoding as a parameterization for the ClusterGen
statistical parametric synthesizer. MCD scores for the three above
described voices are shown in table \ref{Full_build}.

\begin{table}[t,h]
\caption{\label{Full_build}\it{ClusterGen synthesis voice build}}
\centerline{
\begin{tabular}{ |c | c | c| } \hline
  Voice & DNN Params & MCEP params \\\hline
  ARCTIC RMS & 5.851 & 5.161\\
  ARCTIC SLT & 5.466 & 4.858\\ 
  Hindi corpus & 4.680 & 4.134\\ \hline
\end{tabular}}
\end{table}

The Mel Cepstral Distortion is higher for the deep neural network
encoding compared to the default system. In addition to this, the
baseline system was preferred in informal subjective tests. As we had
mentioned earlier in this section, we believe that the MCD of the DNN
systems were affected by the fact that the Deep Neural Network was not
directly optimizing for the score like the default system was
doing. The lack of a good objective metric that would work with the
DNN approach to parameterization exacerbated the problem by making it
difficult to make design decisions. This inturn prevented us from
making use of the full capability of the deep neural network; this is
probably the reason for the lower subjective quality. We believe that
the use of a 
better objective metric would reflect a more positive light on our
results. It would also help us make better decisions which would
contribute towards better parameterizations and improved subjective
results.  

These results are actually quite promising because the relatively good MCD
scores we get with the DNN encoding strongly indicate that the
encoding exists in an \emph{interpolable} space. This is important
because synthesizers like ClusterGen form clusters of the data vectors
at the leaves of the trees and represent the cluster by its
mean\cite{zen2009statistical}. Therefore, only representations like
MCEPs or Line Spectral Pairs\cite{itakura1975line} have been found to
be suitable. The interpolable space constraint is probably the most
difficult to achieve of the four earlier stated constraints. Even
if our data-driven parameterization currently does slightly worse compared to
MCEPs, it is extremely encouraging to be able to find that this
parameterization manages to satisfy all of the four
requirements. Considering how close the difference is between the
performance of the data-driven parameterization and that of the MCEPs,
we expect that a more judicious design of the neural network coupled
with better learning strategies will lead to great results in the
future. 

\subsection{Network optimization}
An investigation of a deep neural network parameterization would be
incomplete without exploring the various possible structures of the
neural network. We tested the neural network both by varying its
width, as well as its depth. Table
\ref{wide_deep} summarizes the results of these experiments.

\begin{table}[t,h]
\caption{\label{wide_deep}\it{Varying the width and depth}}
\centerline{
\begin{tabular}{ |c | c | c| } \hline
  SDA & Analysis- & Voice  \\ 
  Structure & Resynthesis & Build\\\hline
  257 x 125 x 75 x 50             & 4.315 & 5.466 \\
  257 x 750 x 50                  & 4.477 & 5.543\\
  257 x 1000 x 250 x 50           & 4.156 & 5.454\\ 
  257 x 175 x 125 x 75 x 50       & 4.003 & 5.386\\
  257 x 200 x 175 x 125 x 75 x 50 & 3.827 & 5.374 \\ \hline
\end{tabular}}
\end{table}

All of these tests were on the SLT voice. The first column describes
the structure of the Stacked Denoising Autoencoder. The unwrapped Multi Layer
Perceptron that corresponds to each SDA would be twice as deep. 
The network described in the first row of the table is the same
network as the one in the SLT row in table \ref{Full_build}. As
can be seen in the table, for a constant depth, increasing the width
of the layers of the neural network improves the performance of the
network. Increasing the depth, however, makes a much bigger impact on
performance. This is in line with deep learning theory\cite{bengio2011expressive}.  A deep narrow
network also takes substantially less time to train compared to a
shallow wide network.   These trends are encouraging, and we expect further 
investigation in these directions to improve the performance further.

In an earlier section, when we described the training process of a
Denoising Autoencoder, we had mentioned that it is trained to reconstruct
the original input from a noisy version of the input. Traditionally,
this 'noise' that is added involves arbitrarily setting some of the
input parameters to zero. We also investigated using \emph{Gaussian}
noise since this is closer to the noise introduced by statistical
parametric synthesizers. Unfortunately, this only had the effect of making synthesis
quality worse compared to the traditional 'noising' strategy. 

\section{Discussion}

All experiments described in this paper were run on one of the
following Nvidia GPUs: Tesla M2050, GRID GK104, GTX670, GTX660, and
the GTX580.  Finding the right stable hardware combination that offers
the most efficient training platform is also an investigatory task.  

Although we have tried to build on work in related fields (particulary
speech recognition and speech coding) in order to find reasonable
topologies for our networks, the generative nature of speech synthesis
does have inherently different requirements, thus we feel there is
likely significant improvements possible within this core technology.

As it stands, this work concentrates on finding an encoding for
modeling the vocal tract of the speaker, as that allows the most
direct comparison with MCEP parameterization.  However this technology
is in no way constrained by that restriction and adding excitation,
and prosodic information to the networks still fits within our method.

\newpage
\eightpt \bibliographystyle{IEEEtran} \bibliography{refs}

\end{document}